\documentclass[11pt]{article}

\usepackage{fullpage}

\usepackage{microtype}
\usepackage{graphicx}
\usepackage{subfigure}
\usepackage{booktabs} 

\usepackage{hyperref}

\usepackage{amsfonts}
\usepackage{amsthm}
\usepackage{amsmath}
\usepackage{mathtools}
\usepackage{balance}
\usepackage{color}

\newtheorem{theorem}{Theorem}
\newtheorem{lemma}[theorem]{Lemma}
\newtheorem{definition}[theorem]{Definition}

\newcommand{\R}{\mathbb{R}}

\newcommand{\sgn}{\mathrm{sgn}}
\newcommand{\relu}{\mathrm{relu}}
\newcommand{\mean}{\mathrm{mean}}
\newcommand{\std}{\mathrm{std}}
\newcommand{\erf}{\mathrm{erf}}
\newcommand{\sigmoid}{\mathrm{sigmoid}}
\newcommand{\Bern}{\mathrm{Bern}}
\newcommand{\eps}{\epsilon}

\newtheorem{proof-sketch}{Proof-Sketch}
\newtheorem{claim}{Claim}

\newtheorem{observation}{Observation}
\newtheorem{corollary}[theorem]{Corollary}
\newtheorem{conjecture}[theorem]{Conjecture}

\newcommand{\citet}[1]{\cite{#1}}

\newcommand{\mymath}[1]{\[ #1 \]}

\newcommand{\rknote}[1]{}

\begin{document}

\title{On the Learnability of Deep Random Networks}
\author{
Abhimanyu Das
\hspace*{1cm}
Sreenivas Gollapudi
\hspace*{1cm}
Ravi Kumar
\hspace*{1cm}
Rina Panigrahy\\
Google \\
Mountain View, CA
}

\maketitle

\thispagestyle{empty}

\begin{abstract}
In this paper we study the learnability of deep random networks from both theoretical and practical points of view.  On the theoretical front, we show that the learnability of random deep networks with sign activation drops exponentially with its depth.  On the practical front, we find that the learnability drops sharply with depth even with the state-of-the-art training methods, suggesting that our stylized theoretical results are closer to reality.
\end{abstract}

\newpage
\addtocounter{page}{-1}

\section{Introduction}
\label{sec:intro}

Very little is understood about the exact class of functions learnable with deep neural networks.  Note that we are differentiating between representability and learnability.  The former simply means that there exists a deep network that can represent the function to be learned, while the latter means that it can be learned in a reasonable amount of time via a method such as gradient descent.

While several natural functions seem learnable, at least to some degree by deep networks, we do not quite understand the specific properties they exhibit that makes them learnable.  The fact that they can be learned means that they can be represented by a deep network.  Thus, it is important to understand the class of functions that can be represented by deep networks (as \emph{teacher} networks) that are learnable (by any \emph{student} network).

In this work we study the learnability of \emph{random} deep networks. By random, we mean that all weights in the network are drawn i.i.d from a Gaussian distribution.  We theoretically analyze the level of learnability as a function of (teacher network) depth.  Informally, we show the following statement in Section~\ref{sec:sq}.
\begin{quote}
The learnability of a random deep network with $\sgn$ activations drops exponentially with its depth.
\end{quote}
%

Our proof proceeds by arguing that for a random network, a pair of non-collinear inputs progressively repel each other in terms of the angle between them.  Thus, after a small number of layers, they become essentially orthogonal.  While this statement appears intuitively true, proving it formally entails a delicate analysis of a stochastic non-linear recurrence, which forms the technical meat of our work.  Inputs becoming orthogonal after passing through random layers means that their correlation with any fixed function becomes very small.  We use this fact along with known lower bounds in the statistical query (SQ) model to argue that random networks of depth $\omega(\log n)$  cannot be efficiently learned by SQ algorithms~\cite{Kearns98}, which is a rich family that includes widely-used methods such as PCA, gradient descent, etc.  Our proof technique showing non-learnability in the deep network setting could be of independent interest.

While we argue the random deep network becomes harder to learn as the depth exceeds $\omega(\log n)$ where $n$ is the number of inputs, an interesting question is whether these functions are cryptographically hard to learn, i.e., are they cryptographic hash functions?  While we do not definitively answer this question, we provide some evidence that may point in this direction.  Informally we show the following in  Section~\ref{sec:kway}.
\begin{quote}
Outputs of random networks of $\omega(\log n)$-depth are $n^{1/4}$-way independent.
\end{quote}

To confirm that our bounds are not too pessimistic in practice, we take different random deep networks of varying depth and evaluate experimentally to what extent they are learnable using known, state-of-the-art deep learning techniques.  Our experiments show that while these random deep networks are learnable at lower depths, their learnability drops sharply with increasing depth, supporting our theoretical lower bounds.
 
\paragraph{Related work.} 

Several works investigate the learnability of simple networks where the network to be learnt is presented as a blackbox teacher network that is used to train a student network by taking several input-output pairs. If the inputs to the teacher network are drawn from a Gaussian distribution, then for networks of small depth (two) 
~\citet{du2017gradient, du2017convolutional, li2017convergence, zhong2017a, zhang2017electron, zhong2017recovery} guarantee learning the function under certain assumptions.  Our non-learnability results for higher depths are in contrast with these results.

Non-learnability results ~\cite{Daniely16,goel2016reliably,brutzkus2017globally,klivans2009cryptographic} have been shown for depth-two networks assuming adversarial input distributions.  Our work does not assume any adversarial input distribution but rather studies deep networks with random weights.

Another line of research borrows ideas from kernel methods and polynomial approximations to approximate the neural network by a linear function in a high-dimensional space and subsequently learning the same \cite{zhang2015learning, goel2016reliably, goel2017learning, goel2017eigenvalue}. Tensor decomposition methods \cite{anandkumar2016efficient, janzamin2015beating} have also been applied to learning these simple architectures.  On the other hand, while our non-learnability results hold for non-linear activations, we show a matching learnability result with only linear functions.

SQ-based lower bounds have been shown for non-adversarial distribution of inputs for several problems such as learning GMMs and decision trees as~\cite{Bshouty2a,DiakonikolasKS17}.
\section{Background and notation}
\label{sec:prelim}

In this paper we denote vectors by lower case, with subscripts denoting the components; we denote matrices by upper case.  If $f(\cdot)$ is a scalar function and $x$ is a vector, we let $f(x)$ denote the vector obtained by applying $f$ to each component of $x$.  For two vectors $x, y \in \R^n$, let $\theta(x, y)$ denote the angle between $x$ and $y$ and let $x \cdot y$ denote their inner product and let $c(x, y) = \cos(\theta(x, y)) = (x \cdot y)/(\|x\| \|y\|).$
For a random variable $Z$, we denote its mean and standard deviation by $\mean(Z)$ and $\std(Z)$ respectively.  Let $\mathcal{N}(0,1)$ denote the unit normal distribution.

We consider deep networks with the following topology.  There are $n$ inputs to the network, where each input is $\pm 1$; thus the domain of the network is $\{ \pm 1 \}^n$.  The network has depth $h > 1$.  The $i$th layer of the network ($i < h$) operates on an input $x^{(i)}$ and produces the output 
$$x^{(i+1)} = \sigma(W^{(i)} \cdot x^{(i)}),$$
for the next layer, where $\sigma(\cdot)$ is a non-linear activation function and $W^{(i)}$ is an $n \times n$ matrix.  Therefore, each internal layer of the network is fully connected and has the same width%
\footnote{This assumption is not critical for the results; we make this for simplicity of exposition.} as the input.  By convention, $x^{(0)}$ corresponds to the input to the network and the top layer $W^{(h)}$ is a $1 \times n$  matrix so that the final network output $x^{(h+1)}$  is also $\pm 1$.

In this paper the objects of study are random deep networks.  We assume that each $W_i$ is an $n \times n$ random matrix, where each entry is chosen uniformly at random in $[0, 1]$.  Let $f_{W, h, \sigma}(x) = x^{(h+1)}$ be the function computed by a depth-$h$ network with activation $\sigma$ and random weight matrices $W_1, \ldots, W_{h+1}$ on an input $x$.  We will consider  non-linear activation functions such as the sign function ($\sgn(x) = x / |x|$), the rectified linear function ($\relu(x) = \max(0, x)$), and the sigmoid function ($\sigmoid(x) = (1 + \exp(-x))^{-1}$).  

Our analysis will mainly focus on the $\sgn$ activation function; in this case we will abbreviated the computed function as $f_{W, h}(\cdot)$.    The $\sgn$ function satisfies the following well-known elegant geometric property~\cite{GW}: 
\begin{lemma}
\label{lem:hyperplane}
For any two vectors $x, y \in \R^n$, 
$\Pr_{r \in \mathcal{N}(0,1)^n} [\sgn(x \cdot r) \neq \sgn(y \cdot r)] = \theta(x, y)/\pi$.
\end{lemma}
We now define a function that will play an important role in our analysis.
\begin{equation}
\label{eq:sign}
    \mu(x) = \frac{2}{\pi} \arcsin(x).
\end{equation}
This function satisfies the following properties outlined below, which can be seen from the Taylor expansions of $\mu(1-x)$ and $\mu(x)$ respectively.  The first part is an upper bound on the function whereas the second part says that the function can be upper bounded by a linear function for inputs sufficiently close to zero.
\begin{lemma}
\label{lem:useful}
(i) For each $x \in [0, 1]$, $\mu(1-x) \leq 1 - (1/\pi)(\sqrt{2x}).$  (ii) There is a sufficiently small constant $\mu_0$ such that for each $x \in [-\mu_0, \mu_0]$, it holds that 
$|\mu(x)| \leq \rho |x|$ for some constant $\rho = \rho(\mu_0) < 1$.  
\end{lemma}

For $n > 0$ and $p \in [-1, 1]$, let $\Bern(n,p)$ denote distribution of the average of $n$ independent $\pm 1$ random variables such that each random variable is $+1$ with probability $(1+p)/2$ and $-1$ with probability $(1-p)/2$.  Let $c \sim \Bern(n, p)$ denote that the random variable $c$ is distributed as $\Bern(n, p)$.

\section{A repulsion property}
\label{sec:repulsion}

In this section we prove a basic property of random deep networks.  The key point is that whatever be the initial angle between the pair of inputs $x, y$, after a few random $i$ layers, the vectors $x^{(i)}, y^{(i)}$ become close to orthogonal.  Specifically the distribution of the angle becomes indistinguishable from the case had the inputs been chosen randomly.

For the remainder of this section we fix a pair of inputs $x$ and $y$ that are not collinear and study the behavior of a random network on both these inputs.  Let $\theta_i = \theta(x^{(i)}, y^{(i)})$ denote  the angle between the outputs at layer $i$ and let $c_i = \cos(\theta_i)$.  Our goal is to study the evolution of $c_i$ as a function of $i$.

Recall that the $j$th bit of $x^{(i)}$, namely 
$x^{(i)}_j$, is obtained by taking  $\sgn(W^{(i)}_j \cdot x^{(i-1)})$, where $W^{(i)}_j$ is the $j$th row of $W^{(i)}$.  Similarly, 
$y^{(i)}_j = \sgn(W^{(i)}_j \cdot y^{(i-1)})$.
We now calculate the expected value of the product of the bits of $x^{(i)}$ and $y^{(i)}$.
\begin{lemma}
\label{lem:recur}
For any $i$ and any $j \leq n$, it holds that
$E[x^{(i)}_j y^{(i)}_j ~\mid~ c_{i-1}] = \mu(c_{i-1})$.
\end{lemma}
\begin{proof}

Applying Lemma~\ref{lem:hyperplane} to $x^{(i)}$ and $y^{(i)}$ with $W^{(i)}_j$ as the random vector, we obtain
\begin{eqnarray}
\lefteqn{
E[x^{(i)}_j y^{(i)}_j ~\mid~ c_{i-1}] = -1 \cdot \frac{\theta_{i-1}}{\pi} + 1 \cdot \left(1 - \frac{\theta_{i-1}}{\pi} \right)} \nonumber \\
& = & 1 - \frac{2 \theta_{i-1}}{\pi} 
\enspace = \enspace \frac{2}{\pi} \left( \frac{\pi}{2} - \arccos c_{i-1} \right) \nonumber \\
& = & \frac{2}{\pi} \arcsin c_{i-1}
\enspace = \enspace \mu(c_{i-1}). \nonumber
\qedhere
\end{eqnarray}
\end{proof}

We now measure the expected angle between $x^{(i)}$ and $y^{(i)}.$  Note that $c_i \sim \Bern(n, \mu(c_{i-1}))$.
\begin{corollary}
\label{cor:recur}
$E[c_i~\mid~c_{i-1}] = \mu(c_{i-1}).$
\end{corollary}
\begin{proof}
Taking the expectation of
\begin{equation}
\label{eqn:ci}
c_i = \cos(\theta_i) = \frac{ \sum_{j} x^{(i)}_j y^{(i)}_j}{\|x\| \|y\|},
\end{equation}
applying Lemma~\ref{lem:recur}, and using $\|x\| = \|y\| = \sqrt{n}$ finishes the proof.
\end{proof}
To proceed, we next introduce the following definition.
\begin{definition}[$(\epsilon, d)$-mixing]
A pair of inputs $x, y$ is \emph{$(\epsilon, d)$-mixed} if 
$| E[c_d(x, y)] - E[c_d(x', y')]| \leq \epsilon$, where $x', y'$ are chosen uniformly at random in $\{ \pm 1 \}^n$.
\end{definition}
Our goal is to show that $x, y$ is mixed by studying the stochastic iteration defined by Corollary~\ref{cor:recur}.    Note that by symmetry, $E[c_d(x', y')] = 0$.  Hence, to show $(\epsilon, d)$-mixing, it suffices to show $|E[c_d(x, y)]| \leq \epsilon$.  We will in fact show that the iterative process $(\exp(-d), d)$-mixes for $d$ sufficiently large.

\newcommand{\CM}{\mathcal{M}}

\subsection{Mixing using random walks}
\label{sec:mixing}
Corollary~\ref{cor:recur} states a stochastic recurrence for $c_i$; it is a Markov process, where for each $i > 0$, $c_i \sim \Bern(n, \mu(c_{i-1}))$.  However the recurrence is non-linear and hence it is not straightforward to analyze.  Our analysis will consist of two stages.  In the first stage, we will upper bound the number of steps $d$ such that with overwhelming probability $|c_d|$ goes below a sufficiently small constant.  In the second stage, we will condition that $|c_i|, i \geq d$ is below a small constant (assured by the first stage).  At this point, to bound $|E[c_i]|$ we upper bound it by a certain function of the distribution that tracks its level of asymmetry and show that this level of asymmetry drops by a constant factor in each step.  Note that if the distribution is symmetric around the origin then $|E[c_i]| = 0$.

Note that $c_i$ can take at most $2n+1$ distinct values since the inner product between a pair of $\pm 1$ vectors takes only discrete values $-n$ through $n$.  This leads to interpreting the recurrence in Corollary~\ref{cor:recur} as a random walk on a $(2n+1)$-node graph.  (A minor issue is that there are two sink nodes corresponding to $+1$ and $-1$.  We will argue that as long as the inputs are not collinear, the probability of ending up in one of these sink nodes is exponentially small.)

We now begin with the first stage of the analysis.
\begin{lemma}
\label{lem:first}
For $d = O(\log \log n)$ and $\mu_0$ chosen according to Lemma~\ref{lem:useful}, we have
\[
\Pr[|c_d| > \mu_0] \leq d \cdot \exp(-\Omega(\sqrt n)).
\]
\end{lemma}
\begin{proof}
For ease of understanding let us assume $c_0 \ge 0$; an analogous argument works for the negative case.

For convenience, we will track $\epsilon_i = 1 - c_i$ and show that $1 - \epsilon_i$ rapidly goes to a value bounded away from 1.  Initially, 
$\epsilon_0 \geq 1/n$ since the Hamming distance between $x$ and $y$, which are not collinear, is at least $1$.  

From Corollary~\ref{cor:recur} and Lemma~\ref{lem:useful}(i), we have for $i \geq 1$,
\mymath{
E[c_i ~\mid~ c_{i-1}] = \mu(c_{i-1}) = \mu(1 - \epsilon_{i-1}) \leq 1 - \frac{1}{\pi} \sqrt{2\epsilon_{i-1}},
}
equivalently,
\mymath{
E[\epsilon_i ~\mid~ \epsilon_{i-1}] \geq \frac{1}{\pi} \sqrt{2 \epsilon_{i-1}}.
}
This implies by a Chernoff bound that for any $i \geq 1$,
\[
\Pr \left[ \epsilon_i < \frac{1}{2 \pi} \sqrt{2 \epsilon_{i-1}} ~\bigg\vert~ \epsilon_{i-1} \right]
\enspace \geq \enspace e^{-\Omega(E[\epsilon_i ~\mid~ \epsilon_{i-1}] \cdot n)} 
\enspace = \enspace e^{-\Omega(\sqrt{\epsilon_{i-1}} \cdot n)}.
\]
If the above high probability event keeps happening for $d = O(\log \log n)$ steps, then $\epsilon_i$ will keep increasing and
$c_i = 1 - \epsilon_i$ will become less than $\mu_0$ in $d$ steps, where $\mu_0$ is from Lemma~\ref{lem:useful}(ii).  Further, the probability that any of these high probability events do not happen is, by a union bound, at most 
\begin{eqnarray*}
\sum_{i = 1}^d e^{-\Omega(\sqrt{\epsilon_{i-1}} \cdot n)}
& \leq & d \cdot e^{-\Omega(\sqrt{n})},
\end{eqnarray*}
since each $\epsilon_i \geq 1/n$.
\end{proof}

We now move to the second stage.  For the rest of the section we will work with the condition that $|c_i| < \mu_0$ for all $i > d$, where $d$ is given by Lemma~\ref{lem:first}.   First we bound the probability that $c_i$ continues to stay small $h$ steps after $d$.
\begin{lemma}
\label{lem:cont}
For $h > 0$, $\Pr[\exists i \in [d, d+h]: |c_i| > \mu_0 ~\mid~ |c_d| < \mu_0] \leq h e^{-\Omega(n)}$.
\end{lemma}
\begin{proof}
If $|c_{i-1}| \le \mu_0$, $\Pr[|c_i| > \mu_0 ~\mid~ c_{i-1}] \le e^{-\Omega(n)}$ from a Chernoff bound as in the proof of Lemma~\ref{lem:first}.  A union bound finishes the proof.
\end{proof}
From Lemma~\ref{lem:useful}(ii), we know that for $|c| \leq \mu_0$, we have $|\mu(c)| \leq \rho |c|$.

Next we make use of the following Lemma that we prove in the next subsection
\begin{lemma}\label{lemexpdecay}
Conditioning on the absence of the low probability event in Lemma~\ref{lem:cont}, $|E[c_{d+h}]| \le \rho^h$
\end{lemma}

From Lemma~\ref{lem:first} and Lemma~\ref{lemexpdecay}, we 
conclude the following.
\begin{theorem}
\label{thm:repulsion}
For any $x, y$ not collinear we have $|E[c_{d+h}]| \le e^{-\Omega(h)} + h e^{-\Omega(n)} + d e^{-\Omega(\sqrt n)} $. 
\end{theorem}

\rknote{Account for sink nodes}

\subsection{Exponential decay}

We will now show Lemma~\ref{lemexpdecay} that $|E[c_i]|$ decays exponentially in $i$. For ease of exposition throughout this subsection we will implicitly condition on the event that $|c_i| < \mu_0$ right from the initial step through
the duration $h$ of the execution of the Markov process (since probability of this happening is exponentially small, it alters the bounds only negligibly).

For convenience, let $p_c(j)$ denote $\Pr[c=j]$. Define
\[
\Phi_c = \sum_{j>0} j \cdot |p_c(j) - p_c(-j)|.
\]
Note that if $p_c$ is an even function (i.e., the distribution $p_c(\cdot)$ is symmetric around the origin), then we have $\Phi_c = 0$. 
We will show that
\begin{lemma}
\label{lem:phistep}
$\Phi_{c_{i+1}} \le \rho \Phi_{c_{i}}$.
\end{lemma}
\begin{proof}
We first show this statement holds for all point-mass distributions.  Suppose $c_i$ has support only at point $j$.  Recall that $c_{i+1}$ is obtained by first computing $\mu(c_i)$ and then taking the distribution $\Bern(n,\mu(c_i))$.  By the conditioning that $|c_i| < \mu_0$ and from Lemma~\ref{lem:useful}, we have $|\mu(c_i)| \le \rho |c_i|$. Without loss of generality assume $\mu(c_i) \ge 0$. Then for $j \ge 0$, $p_{c_{i+1}}(-j) \le p_{c_{i+1}}(j)$ since this can be viewed as a biased coin that has a higher probability of heads than tails and therefore the probability of getting $j$ more heads than tails is more than the converse probability. 
So 
\[
\Phi_{c_{i+1}} = \sum_{j> 0} j|p_{c_{i+1}}(j) - p_{c_{i+1}}(-j)| = \sum_{j> 0} j p_{c_{i+1}}(j) - j p_{c_{i+1}}(-j) = E[c_{i+1}] = \mu(c_i) 
\le \rho c_i
= \rho \Phi_{c_{i}}, 
\]
where the last step uses $\Phi_{c_{i}} = c_i$ for point-mass.

Note that $\Phi$ when viewed as a function of the distribution $p_c$ is a convex function (since the absolute value operator is convex).

Next we use the statement for point-mass distributions to derive the same conclusion for two-point distributions with support at $\pm j$.  Any such distribution is a convex combination of two distributions: one of them is symmetric and is supported on both points with equal mass and the residual is supported only at either $j$ or $-j$ (whichever had a higher probability).  The former distribution is symmetric and hence when taken to the next step will result again in a symmetric distribution which, as observed earlier, must have $\Phi$ value of $0$. The latter initially has $\Phi$ value equal to $ \Phi_{c_i}$ and after one step will have $\Phi$ value at most $\rho  \Phi_{c_i}$ by our conclusion for point-mass distributions.  The  conclusion for two-point distributions then follows from the convexity of $\Phi$.

Finally, we use these steps to derive the same conclusion for all distributions.  
By the convexity of $\Phi$, any distribution $c_i$ can be expressed as a convex linear combination of two-point distributions $c_{i,1}, \ldots, c_{i,n}$. The distribution $c_{i+1}$ is also the same convex combination of the resulting distributions $c_{i+1,1}, \ldots, c_{i+1,n}$. 
Now if $p_{c_i} = \sum_j \alpha_j c_{i,j}$ then  $\Phi_{c_i} = \sum_j \alpha_j \Phi_{c_{i,j}}$.
Now $p_{c_{i+1}} = \sum_j \alpha_j p_{c_{{i+1},j}}$. So by the convexity of the $\Phi$ function, we have
\[
\Phi_{c_{i+1}} 
\le \sum_j \alpha_j \Phi_{c_{{i+1},j}} 
\le \rho \sum_j \alpha_j \Phi_{c_{{i},j}} = \rho \Phi_{c_i}.
\qedhere
\]
\end{proof}
This immediately yields the following.
\begin{corollary}
$|E[c_i]| \le \rho^i$
\end{corollary}
\begin{proof}
Note that
\[
E[c_i] = \sum_j p_{c_i}(j) j = \sum_{j> 0} j(p_{c_i}(j) - p_{c_i}(-j)) \le  \sum_{j> 0} j|p_{c_i}(j) - p_{c_i}(-j)| \le \Phi_{c_i}.
\]
Applying Lemma~\ref{lem:phistep} inductively completes the proof.
\end{proof}

\newcommand{\STAT}{{\sf STAT}}
\newcommand{\SQ}{\mathsf{SQ}}
\newcommand{\CF}{\mathcal{F}}

\section{Learnability}
\label{sec:learnability}

In this section we discuss the implications of the repulsion property we showed in Section~\ref{sec:repulsion}.  We first calculate the statistical correlation of the function computed by a random network with some fixed function.  We use the correlation bound to argue that random networks are hard to learn in the statistical query model.  Finally we show an almost matching converse to the hardness result:  we show upper bounds on the learnability of a random network.

\subsection{Statistical correlations}

We will now show that the statistical correlation $f_{W, h}(\cdot)$, for random $W$, with any fixed function $g(\cdot)$ is exponentially small in the depth $h$ of the network.  Consider all the inputs $x \in \{ \pm \}^n$ that are in one halfspace; this ensures that no two inputs are collinear.   Note that the number of such inputs is $N = 2^{n-1}$.  Define for $p \geq 1$, the norm $|g|_p = E_x[|g(x)|^p]^{1/p}$. 
\begin{lemma}
\label{lem:corr}
$E_W [ E_x[g(x) f_{W, h}(x)]^2 ] 
 \le |g|_2^2/N + |g|_1^2 e^{-h}$. 
\end{lemma}
\begin{proof}
\begin{eqnarray*}
\lefteqn{
E_W [ E_x[g(x) f_W(x)]^2 ]
} \\
& = & \frac{1}{N^2} \left( \sum_x E_W[g(x)^2 f_W(x)^2] \right. \\
& & \quad \left. + 2 \sum_{x \neq y} E_W[g(x) g(y) f_{W, h}(x) f_{W, h}(y)] \right)
\\
& = & \frac{1}{N^2} \sum_x g(x)^2  + \frac{2}{N^2} \sum_{x \neq y} g(x) g(y) E_W [f_{W, h}(x) f_{w, h}(y)]
\\
& \le & \frac{1}{N} E_x[g(x)^2] + e^{-h} \frac{2}{N^2} \sum_{x \neq y} |g(x)| |g(y)| 
\\
& \le & \frac{1}{N}  |g|_2^2   + e^{-h} |g|_1^2,
\end{eqnarray*}
using Theorem~\ref{thm:repulsion}.
Note that this bound is $e^{-\Omega(h)}$ when $g$ has bounded first and second norms and when $h = o(\sqrt{n})$.
\end{proof}

\subsection{Statistical query hardness}
\label{sec:sq}

In this section we show how the statistical correlation bound we obtained in Lemma~\ref{lem:corr} leads to hardness of learning in the statistical query (SQ) model.  The SQ model captures many classes of powerful machine learning algorithms including gradient descent, PCA, etc; see for example the survey by \citet{vitaly}.

We briefly recap the \emph{SQ model}~\cite{Kearns98,Bshouty2a}.  In this model, instead of working with labeled examples, a learning algorithm can estimate probabilities involving the target concept.  Let $X$ denote the space of inputs.  Formally speaking, a statistical query oracle $\SQ_{f, D}(\psi, r)$ for a Boolean function $f: X \rightarrow \{\pm 1\}$ with respect to a distribution $D$ on $X$ takes as input a query function $\psi : X \times \{-1, +1\} \rightarrow [-1, +1]$ and a tolerance $r$ and returns $v$ such that $|E_D[\psi(x, f(x))] - v| \leq r$.  A concept class $\CF$ is \emph{SQ-learnable} with respect to a distribution $D$ if there is an algorithm $\mathcal{A}$ that, for every $f \in \CF$, obtains an $\epsilon$-approximation%
\footnote{
A function $f$ is an \emph{$\epsilon$-approximation} to $g$ with respect to $D$ if $\Pr_D[f(x) = g(x)] \geq 1 - \epsilon$.
}
to $f$ with respect to $D$ in time $p(n, s, 1/\epsilon)$ for some polynomial $p(\cdot)$.  Here $n$ is the number of parameters to describe any $x \in X$ and $s$ is the number of parameters to describe any $f \in \CF$.  We assume that the evaluation of the function that $\mathcal{A}$ uses for the $\SQ_{f, D}$ oracle takes time $q(n, s, 1/\epsilon)$ and the tolerance is $1/r(n, s, 1/\epsilon)$, where $q(\cdot)$ and $r(\cdot)$ are also polynomials.

It has been shown that for Boolean functions, the statistical query function $\psi$ can be restricted to be a \emph{correlation query} function of the form $\psi(x, f(x)) = g(x) f(x)$, where $g(x) \in [-1, 1]$.  The following can be inferred from Lemma~12 and the arguments in Section~5.1 in \citet{Bshouty2a}.
\begin{lemma}
\label{lem:sq2corr}
For Boolean function $f$, all statistical queries made by an SQ-learning algorithm can be restricted to be correlational queries.
\end{lemma}
%
%
\begin{theorem}[\citet{Bshouty2a}, Theorem~31]
\label{thm:vitaly}
Let $\CF$ be a concept class that is SQ-learnable with respect to $D$, particularly, there exists an algorithm $\mathcal{A}$ that for every $f \in \CF$, uses at most $p(n, s)$ queries of tolerance lower bounded by $1/r(n,s)$ to produce a $(1/2 - 1/q(n,s))$-approximation to $f$. Let $r'(n, s) = \max\{ 2r(n, s), q(n, s) \}$.  There exists a collection of sets $\{ W_i  \}_{i = 1}^{\infty}$ such that $|W_s| \leq p(n, s)+1$ and the set $W_s$ $(1/2 - 1/r'(n,s))$-approximates%
\footnote{
A set $W$ \emph{$\epsilon$–approximates} $F$ with respect to distribution $D$ if for every $f \in F$ there exists $g \in W$ such that $g$ $\epsilon$-approximates $f$ with respect to $D$.
}
$\CF$ with respect to $D$.
\end{theorem}
Using these, we now show the hardness result.
\begin{theorem}
\label{thm:hardness}
If  $E_{f \in \CF} [ E_D [g(x) f(x)]^2 ] < \epsilon(n, s)$, then the class $F^s$ of functions is not SQ-learnable unless $p(n, s) \cdot \max(r(n, s), q(n, s))^2 > 1/\epsilon(n, s)$.
\end{theorem}
\begin{proof}
\rknote{This proof needs rewrite}
Given that the square of the correlation of a uniformly chosen random function $f \in \CF$ is at most $\epsilon$, by a Markov bound, with probability $1 - 1/p(n, s)$, the correlation with a fixed $g$ will also be at most $p(n, s)  \cdot \epsilon(n, s)$.   By a union bound, the correlation squared with all the functions in $W_s$ will then be at most $O(p(n, s) \cdot \epsilon(n, s))$;  i.e., the absolute value of the correlation is $O(\sqrt{p(n, s) \cdot \epsilon(n, s)})$.  However, for the desired tolerance and learnability in Theorem~\ref{thm:vitaly}, there is a function $g \in W_s$ such that $g$ $(1/2 - 1/r'(n, s)$-approximates $f$, i.e., the correlation of $g$ with $f$ is $1/r'(n, s)$.  Thus, we need
$O(\sqrt{p(n, s) \cdot \epsilon(n, s)}) \geq 1/r'(n, s)$, which completes the proof.
\end{proof}
In our case, will use $h$ instead of $s$ to parameterize the target class.  Since $g(x) \in [-1, 1]$, $|g|_1 \leq 1$ and the following SQ lower bound follows then follows from Lemma~\ref{lem:corr} and Theorem~\ref{thm:hardness}.

\begin{theorem}
\label{thm:sqhardnessdeepnetwork}
For any SQ algorithm for learning a random deep network of depth $h$
$p(n, h) \cdot \max(r(n, h), q(n, h))^2 > e^{O(h)}$. In particular this means that to get accuracy that is a constant better than $1/2$, the number of queries $p$ is exponential in $h$; If the number of queries is $poly(h)$ then the improvement over an accuracy of $1/2$ is $exp(-h)$.
\end{theorem}

In the next section, we will show a matching upper bound of $exp(-h)$  for the achievable accuracy in polynomial time.
\subsection{An upper bound}
\label{sec:lb}

We will now show that $f_{W, h, \sigma}$, i.e., a random network of depth $h$ with $\sigma$ activation, can indeed be learned up to an accuracy of $e^{-O(h)}$, in fact, by a linear function.  This essentially complements the hardness result in Section~\ref{sec:sq}.  

We will show that the linear function defined by $g_{W, h}(x) = W^{(h+1)} W^{(h)} \cdots W^{(1)} x$ is highly correlated with $f_{W, h, \sigma}(x)$. We complete the proof by induction.
\begin{lemma}
$E[g_{W, h}(x) \cdot f_{W, h, \sigma}(x)] = e^{-O(h)}$.
\end{lemma}
\begin{proof}
\rknote{Does this work for all $\sigma$?}
We study the output at each layer for both $f$ and $g$.  Let $y$ (resp., $z$) denote the vector of inputs to the $h$th layer of $f_{W, h, \sigma}(x)$ (resp., $g_{W, h}(x)$).  Let $\hat{y} = y / \|y\|$ and $\hat{z} = z / \|z\|$; note that $\|y\| = \sqrt{n}$ for $\sigma = \sgn$.  If $\theta = \theta(y, z)$, then we have 
$\cos(\theta) = \hat{y} \cdot \hat{z}$.  Now,
\begin{eqnarray}
\lefteqn{
\mkern-36mu E_W[f_{W, h, \sigma}(x) \cdot g_{W, h}(x)] = 
\sum_i E[\sigma(W^{(h)}_i \cdot y) (W^{(h)}_i \cdot z)]} \nonumber \\
& = & \|z\| \sum_i E_W[\sigma(W^{(h)}_i \cdot y) (W^{(h)}_i \cdot \hat{z})].  \label{eq:ub1}
\end{eqnarray}
The key step is to express $\hat{z} = \hat{y} \cos\theta + y_{\perp} \sin\theta$.  Fix an $i$ and for brevity let $w = W^{(h)}_i$.  Since $w \sim N(0,1/n)^n$,  we obtain
\begin{eqnarray}
\lefteqn{
E_w[\sigma(w \cdot y) (w \cdot \hat{z})]} \nonumber \\
&  = & E[\sigma(w \cdot y) \left(w \cdot \hat{y} \cos\theta +  w \cdot y_{\perp} \sin \theta \right)] \nonumber \\
& = & E[\sigma(w \cdot y)(w \cdot \hat{y}) \cos\theta] 
+  E[\sigma(w \cdot y) (w \cdot y_{\perp}) \sin \theta] \nonumber \\
& = & E[\sigma(w \cdot y)(w \cdot \hat{y})] \cos\theta 
+  E[\sigma(w \cdot y)] E[w \cdot y_{\perp}] \sin \theta \nonumber \\
& = &\frac{\cos \theta}{\sqrt{n}} E[\sigma(w \cdot y)(w \cdot y)] 
+  E[\sigma(w \cdot y)] E[w \cdot y_{\perp}] \sin \theta \nonumber \\
& = & \frac{\gamma}{\sqrt{n}} \cos \theta. \label{eq:ub2}
\end{eqnarray}
Here, the third equality follows since the random variables $w \cdot y$ and $w \cdot y_{\perp}$ are independent.  The fourth equality follows since $\|y\| = \sqrt{n}$.  The final equality follows since (i) $w \cdot y \sim \|y\| N(0, 1/n) \sim N(0, 1)$ and hence the expectation in the first term is of the form $E_r[\sigma(r) \cdot r]$ where $r \sim  N(0, 1)$, which is some absolute constant%
\footnote{
For $\sigma = \sgn$, $\gamma$ is the mean of the folded unit normal distribution and is  $\sqrt{2/\pi} \approx 0.8$.
} $\gamma < 1$ that depends on $\sigma$
and
(ii) the second term vanishes since 
$w \cdot y_{\perp} \sim \| y_{\perp} \| N(0, 1/n)$.
Plugging (\ref{eq:ub2}) into (\ref{eq:ub1}), we obtain
\begin{eqnarray*}
\lefteqn{
E_W[f_{W, h, \sigma}(x) \cdot g_{W, h}(x)] = \|z\| \cdot n \cdot \frac{\gamma}{\sqrt{n}} \cos \theta} \\
& = &  \|y\| \|z\| \cdot \gamma \cos \theta 
\enspace = \enspace \|y\| \|z\| \cdot \gamma (\hat{y} \cdot \hat{z}) \\
& = & \gamma (y \cdot z) 
\enspace = \enspace \gamma^h n,
\end{eqnarray*}
by an induction on $h$ and using the base case at the input layer where $x \cdot x = n$.  
Since the final output of $f_{W, h, \sigma}$ is a bit, the average correlation with $g_{W, h}$ is $\gamma^h = e^{-O(h)}$.
\end{proof}

\section{On extending to $\relu$}
\label{sec:relu}

We now extend some of the previous analysis of the $\sgn$ activation function to the $\relu$ activation function. As we describe later in Section \ref{sec:expts} for the $\relu$ (and $\sigmoid$) teacher networks, to keep the output vectors at each layer bounded, we normalize the output of each layer to have mean $0$ and variance $1$ along each dimension. This is analogous to the batch normalization operation commonly used in training; in our case, we refer to this operator as $B(\cdot)$. Thus, the operation at each layer is the following: (i) multiply the previous layer's outputs by a random Gaussian matrix $W$, (ii) apply $\relu$, and (iii) apply the normalization operation $B$.

As in the previous analysis, the key step is to show that whatever be the initial angle between the pair of inputs $x, y$, after a few random $i$ layers with $\relu$, the vectors $x^{(i)}, y^{(i)}$ quickly become close to orthogonal.

We fix a pair of input vectors $x_i$ and $y_i$ with length $\sqrt{n}$ and inner product $c_0$ and study the behavior of this $\relu$ layer on the inner product $c_1$ of the corresponding output vectors.  Recall that $c(x, y) = \cos\theta(x, y)$.  
In particular, we are interested in $E_W[c_{i+1} ~\mid~ c_i] = E_W[c(x_{i+_1}, y_{i+1})] = E_W[c(B(\relu(W x_i),  B(\relu(W y_i))]$, where the batch normalization operator for $\relu$ is the function $B(x) = \frac{x - \mean_{Z \in \mathcal{N}(0,1)}(\relu(Z))}{\std_{Z \in \mathcal{N}(0,1)} (\relu(Z))}$.

 By 2-stability of Gaussians and since $W$ is a random  matrix, $W x_i$ and $W y_i$ are also Gaussian random vectors. Note that we do not not make any assumption that input vectors at each layer are Gaussian distributed: $x_i$ and $y_i$ are arbitrary fixed vectors, however multiplying them with a random $W$ matrix leads to Gaussian random vectors. We get the following (where we use $x$ and $y$ to refer to $W x_i$ and $W y_i$ respectively).
\begin{lemma}
$E_W[c_{i+1} ~\mid~ c_i] = 
E_{x,y \in \mathcal{N}(0,1)^n, c(x,y) = c_i}[c(B(\relu(x)),  B(\relu(y)))].$
\end{lemma}

Thus, we need to analyze how the angle between two random Gaussian vectors changes when passed through a $\relu$ and the normalization operator $B(\cdot)$. 
In particular, given two vectors $x, y \in \mathcal{N}(0,1)^n$ at angle $\theta$ and cosine similarity $c_0=\cos(\theta)$,  we wish to find the expected cosine similarity ($c_1$) after applying one iteration. Furthermore, since all dimensions of $x$ and $y$ are independent, using an argument similar to that in Corollary~\ref{cor:recur} we can consider the expected dot product along only one dimension of the Gaussian vectors. Thus, we have %
\begin{corollary}
$E_W[c_{i+1} ~\mid~ c_i] = 
E_{x,y \in \mathcal{N}(0,1), x.y = c_i}[B(\relu(x)) \cdot B(\relu(y))]$. 
\end{corollary}

\begin{figure}
\centering
\includegraphics[angle=-90,width=3in]{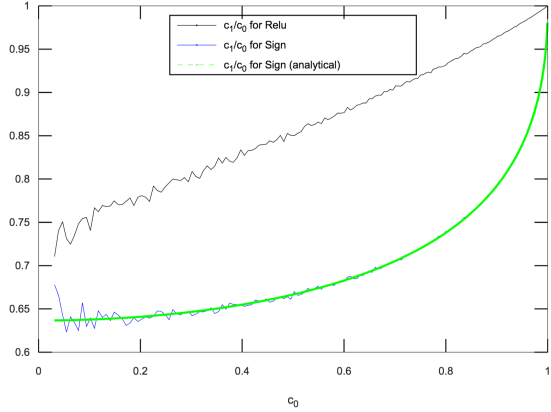}
\caption{Ratio of $c_1/c_0$ for $\relu$ and $\sgn$.  \label{figure:relu}}
\end{figure}

We now make the following claim (proof in Appendix).
\begin{lemma}
\label{lem:relu1}
$|E_{x,y \in \mathcal{N}(0,1), c(x,y) =c_0} [B(\relu(x)) \cdot B(\relu(y))]| \leq |c_0|.$
\end{lemma}
In Figure~\ref{figure:relu} we plot the ratio of this expected value (which we denote as $c_1$) and $c_0$  via simulations, which shows that this ratio is always less than 1 for all values of $c_0$. In addition to the plot for $\relu$, we also plot the ratio $c_1/c_0$ for $\sgn$ function (obtained via simulations and via its closed-form expression given by (\ref{eq:sign}).

However, we are unable to account for the distribution around the expected value and prove that the magnitude of the expected value of the sum of $n$ such random variables drops exponentially with height.  Hence, we leave this as a conjecture that we believe to be true.

\begin{conjecture}
For any $x, y$ not collinear we have $|E[c_{h}]| \le e^{-\Omega(h)}$ for all $h > 0$.
\end{conjecture}

As in the $\sgn$ case, SQ-hardness hardness for $\relu$ activation follows directly from the above conjecture, using Section~\ref{sec:learnability}.
\section{$k$-way independence}
\label{sec:kway}

\begin{definition}
$\eps$-additive-close to full independence: A set of $k$ bits is independent if the probability of seeing a specific combination of values is  $1/2^k$. It is $\eps$ close to fully independent if its probability distribution differs from the fully independent distribution in statistical distance by at most $\eps$.
\end{definition}

\begin{definition}
$\eps$-additive-close to   $k$-way independence: A binary function is $\eps$-additive-close to $k$-way independent if the outputs any $k$ inputs is $\eps$-additive-close to fully independent.
\end{definition}

\begin{definition}
$(1 \pm \eps)$-multiplicative-close to full independence: A set of $k$ bits $1 \pm \eps$ factor close to independent if the probability of seeing a specific combination of values is  $(1 \pm \eps).1/2^k$.
\end{definition}

Now take a subset of $k$ binary inputs $x$ so that no pair is equal to one another or opposite; that is, $x_i \ne \pm x_j$.
We will also exclude opposites among the $k$ inputs.

\begin{theorem}
The function $f$ with depth $\Omega (\log n)$ is $\tilde O(k^2/\sqrt{n})$-additive close to $k$-way-independent
\end{theorem}

\begin{theorem}
For a function $f$ of depth $d = \Omega (\log n)$, with high probability of $1-1/poly(n)$ over the first $d-1$ layers, the output after the last layer is $e^{\pm \tilde O(k^2/\sqrt{n})}$-multiplicative close to fully independent. 
\end{theorem}

\begin{observation}
If $k$ outputs at a layer $i$ are orthogonal then after one layer the outputs $y_{i+1}$ form a set of $k$ fully independent $n$ dimensional vectors.
\end{observation}

The main idea behind the following lemma is that the $k$ outputs become near orthogonal after $\Omega(\log n)$ layers. 

\begin{lemma}
After $h=\Omega(\log n)$ steps of the random walk $|c_h| = O(\sqrt{s/n})$ with probability $1-h/s$ 
\end{lemma}
\begin{proof}
We already know that $|\mu(c_i)|$ becomes less than some constant after $O(\log\log n)$ steps and after that it decreases by factor $\rho$ in expectation in each iteration. The sampling from $Bern(n,\mu(c_i))$ can cause a perturbation of at most $O(\sqrt{s/n})$ w.h.p. $1-1/s$. So w.h.p. $|c_i| \le \rho |c_{i-1} + O(\sqrt{s/n})$. Chasing this recursion gives that after $h=\Omega(\log n)$ steps, $|c_h| = O(\sqrt{s/n})$ w.h.p. $1 - h/s$.
\end{proof}

We say that a pair of vectors $x_i,x_j$ are near $\delta$-near orthogonal if normalized dot products $|x_i \cdot x_j|/(\|x_i\| \|x_j\|) \le \delta$

\begin{corollary}
For $k$ inputs $x_1,..,x_k$, w.h.p. after $O(\log n)$ layers, all pairs from the set of $k$ inputs are pairwise  $O(\sqrt{s/n})$-near orthogonal w.h.p. $1- k^2h/s$
\end{corollary}

\begin{claim}
If $k$ inputs to a layer $x_1,..,x_k$  are $\delta$-near orthogonal then after one layer the $k$ outputs at any node at that next layer is $e^{\pm \delta k^2}$-multiplicative close to independent.
\end{claim}
\begin{proof}
We are dotting the $k$ inputs with a random vector $w$ and then taking sign. If $X$ is the matrix formed by these inputs as rows $x_i^T$, then consider the distribution of $y = X w$. If rows of $X$ were orthogonal, $y$ is a uniform Gaussian random vector in $R^k$. If $X$ is near orthogonal, it will depart slightly from a uniform Gaussian. Consider the pdf.

$y$ is also $k$ dimensional normally distributed vector with $k \times k$ covariance matrix $U = E_w[Xww^TXT] = XX^T/n$.

So pdf of $y$ will be $c.\frac{1}{\sqrt{\det(U)}} e^{-x'U^{-1}x}$ for some constant $c$. If all the normalized dot products are at most $\delta$. Then $U = I + \Delta$ where each entry in $\Delta$ is at most $\pm \delta$. So eigenvalues of $\Delta$ is at most $\pm \delta k$  and of $U$ is within $1 \pm \delta k$. So the determinant is $e^{\pm O(\delta k^2})$. So over all the distribution differs from that of a uniform normal distribution by $e^{\pm O(\delta k^2)}$.
\end{proof}
Since w.p. $1 - 1/n^{O(1)}$, the normalized dot product is $\pm O(\sqrt{\log n / n})$, we know that conditioned on this, we get a $1+O(k^2 \sqrt{\log n/n})$-multiplicative close to fully independent outputs. So over all it is $O(k^2 \sqrt{\log n / n})$-additive-close to fully independent. This gives.

\begin{theorem}
The function $f$ after depth $\Omega(\log n)$ is $O(k^2 \sqrt{\log n / n})$-additive close to $k$-way-independent 
\end{theorem}

\section{Conclusions}
\label{sec:conc}

We considered the problem of learnability of functions generated by deep \emph{random} neural networks. We show theoretically that deep random networks with $\sgn$ activations are hard to learn in the SQ model as their depth exceeds $\omega(\log n)$, where $n$ is the input dimension.  Experiments show that functions generated by deep teacher networks (with $\sgn, \relu, \sigmoid$ activations) are not learnable by any reasonable deep student network, beyond a certain depth, even using the state-of-the-art training methods.  Our work motivates the need to better understand the class of functions representable by deep teacher networks that make them amenable to being learnable.

\section*{Appendix}

\appendix

\newcommand{\teacher}{\mathrm{teacher}}
\newcommand{\student}{\mathrm{student}}

\section{Experiments}
\label{sec:expts}

We next describe experiments that support our theoretical results and show that random networks using $\sgn, \relu$ or $\sigmoid$ activations are not learnable at higher depths.

\paragraph{Experimental setup and data.}

Since we wish to evaluate learnability of deep random networks, we rely on generating synthetic datasets for binary classification, where we can ensure that the training and evaluation dataset is explicitly generated by a deep neural network. In particular, we specify a \emph{teacher network} parameterized by a choice of network depth $d$, layer width $w$, activation function $\sigma$, and input dimension $n$. Each layer of the teacher network is a fully connected layer specified by the width and activation function. The last layer is a  linear layer that outputs a single float, which is then thresholded to obtain a Boolean label. We initialize all parameters in the network with random Gaussian values. In particular, we initialize all the matrices $W^{(1)}, W^{(2)}, \ldots, W^{(d)}$.  

In addition, as mentioned in Section~\ref{sec:relu}, we re-normalize the output of each layer to have mean $0$ and variance $1$ along each dimension (since otherwise the function becomes trivially learnable as most outputs concentrate around $0$ with increasing depth). We generate $10$ million training examples for each teacher network. For each training example, we generate a uniform random input vector in the range $[-0.5,0.5]$ as an input to this network, and use the Boolean output of the network to generate (positive or negative) labels.
We generate different teacher networks by varying the activation functions ($\sgn$, $\relu$, and $\sigmoid$) and depth (from $2$ to $20$) for the network. We fix the width at $50$ and input dimension at $512$.

We try to learn the classification function represented by each teacher network, with a corresponding \emph{student} network which is assumed to know the topology and activation function used by the teacher network, but not the choice of weights and biases in the network. Further, we use student networks for varying depths $\ge h$ and widths $ \ge w$. For the case of a teacher network with $\sgn$, since the activation function is not differentiable, we use a $\relu$ function in the corresponding student network. We configure dropout, batch-normalization, and residual/skip-connections at each layer of the teacher network, and tune all hyperparameters before reporting our experimental results.

\paragraph{Results.}

\begin{figure*}[ht]
  \subfigure[$\sigma = \sgn$] {
    \centering
    \includegraphics[width=0.27\textwidth]{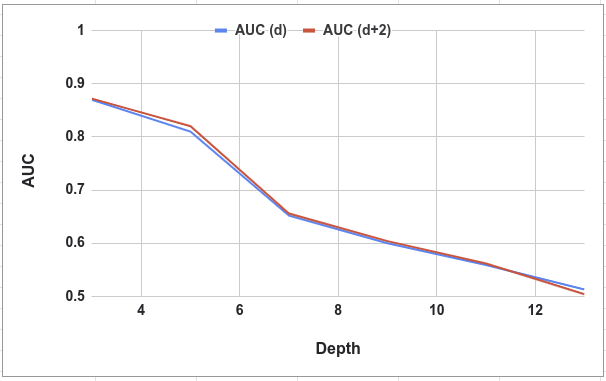}
    \label{fig:cr}
  }
  \qquad
  \subfigure[$\sigma = \relu$] {
    \centering
    \includegraphics[width=0.27\textwidth]{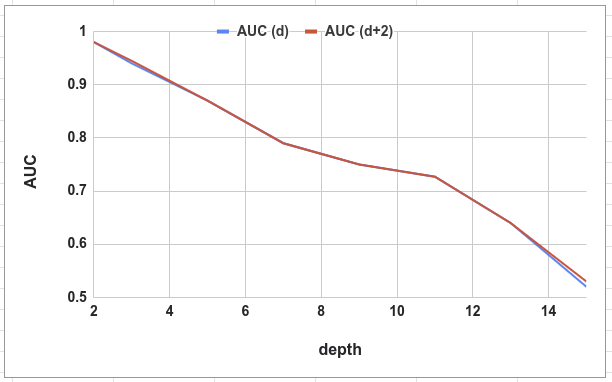}
    \label{fig:rvc}
  }
  \qquad
  \subfigure[$\sigma = \sigmoid$] {
    \centering
    \includegraphics[width=0.27\textwidth]{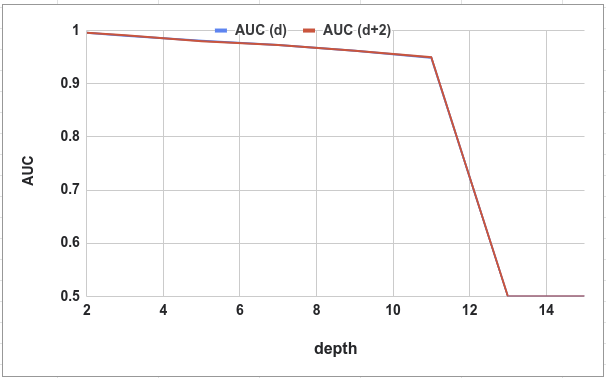}
    \label{fig:hybrid}
  }
  \caption{The learnability of teacher networks with different activation functions $\sigma$.}
  \label{fig:expts}
\end{figure*}

Figure~\ref{fig:expts} illustrates the learnability of the functions generated by the teacher network for the different parameters mentioned above. In particular, we report results for  $\sigma \in \{\sgn, \relu, \sigmoid \}$, $d_{\student} \in \{d_{\teacher}, d_{\teacher}+2\}$, $w_{\student} = w_{\teacher}$. We observe that for all three activation functions, the test AUC of the student network falls starts close to $1$ at small depths and drops of rapidly to $0.5$ as we increase the depth. This is consistent with our theoretical results summarized in Section~\ref{sec:learnability}. While this behavior is more representative in the case of $\sgn$ and $\relu$ teacher networks, the $\sigmoid$ activation function seems to exhibit a threshold behavior in the AUC. In particular, even as the AUC stays as high as $0.95$ for $d = 11$, it drops sharply to $0.5$ for greater depths. We leave the analysis of this intriguing behavior as an open problem.

In another experiment, we studied the effect of overparameterizing the width and height parameters of the student network when compared to the teacher network. To this end, we varied the width of the student network as $[w_{\teacher}, 2w_{\teacher}$ and the depth as $[d_{\teacher}, d_{\teacher}+6]$.  We did not observe any significant change in the learnability results for all combinations of these parameter values.

\rknote{How many samples}
\section{Proof of Lemma~\ref{lem:relu1}}

\begin{proof}
Without loss of generality, let $y = c_0 z + \sqrt{1-c_0^2} z$, such that $z \in \mathcal{N}(0,1)$ and $cov(x,z)= 0$. This implies that $p(x,z) = p(x) p(z)$.

We will first bound $E_{x,y}[\relu(x)) \cdot \relu(y)]$ (ignoring the Batch Normalization operator).

We have
\begin{eqnarray*}
E_{x,y}[\relu(x)) \cdot \relu(y)] & = & \int_x \int_z \max(0,x) \max(0, c_0 x + \sqrt{1-c_0^2} z) p(x,z) dx dz \\
& = & \int_{x=0}^\infty x\int_{z=\frac{-c_0 x}{\sqrt{1 - c_0^2}}}^\infty (x c_0 + \sqrt{1-c_0^2} z)  p(x,z) dx dz \\
& = &  \int_{x=0}^\infty  x^2 c_0 p(x) dx \int_{z=\frac{-c_0 x}{\sqrt{1 - c_0^2}}}^\infty p(z) dz + \int_{x=0}^\infty  x p(x)dx \int_{z=\frac{-c_0 x}{\sqrt{1 - c_0^2}}}^\infty \sqrt{1 - c_0^2} z p(z)dz \\
& = & \int_{x=0}^\infty  x^2 c_0 p(x) dx \int_{z=0}^\infty p(z) dz + \int_{x=0}^\infty  x^2 c_0 p(x) dx \int_{z=0}^{\frac{c_0 x}{\sqrt{1 - c_0^2}}} p(z) dz + \\
& & \int_{x=0}^\infty  x p(x)dx \int_{z=\frac{-c_0 x}{\sqrt{1 - c_0^2}}}^\infty \sqrt{1 - c_0^2} z p(z)dz \\
& = & \int_{x=0}^\infty \frac{ x^2 c_0}{\sqrt{2 \pi}} p(x) dx \int_{z=0}^\infty e^{-z^2/2} dz + \int_{x=0}^\infty  \frac{ x^2 c_0}{\sqrt{2 \pi}} p(x) dx \int_{z=0}^{\frac{c_0 x}{\sqrt{1 - c_0^2}}} e^{-z^2/2} dz + \\
& & \int_{x=0}^\infty  \frac {x \sqrt{1 - c_0^2}}{\sqrt{2 \pi}} p(x)dx \int_{z=\frac{-c_0 x}{\sqrt{1 - c_0^2}}}^\infty z e^{-z^2/2} dz \\
\end{eqnarray*}

We make use of the fact that $\int z e^{-z^2/2} dz = -e^{-z^2/2}$ and that the definite Gaussian integral $\int_{0}^\infty e^{-z^2/2} = \sqrt{2\pi}/2$. We also use the Error function definition $\erf(x) = \frac{2}{\sqrt{\pi}} \int_{u=0}^x e^{-u^2}du$, and substitute variables $x$ by $v=\frac{c_0 x}{\sqrt{2(1- c_0^2)}}$ in the second integration term. We therefore get  
\begin{eqnarray*}
E_{x,y}[\relu(x)) \cdot \relu(y)] & = & \int_{x=0}^\infty \frac{1}{2} x^2 c_0 p(x) dx + \int_{v=0}^\infty \frac{(1 - c_0^2)^{1.5}}{c_0^2 \sqrt{\pi}} e^{-\frac{v^2 (1 - c_0^2}{c_0^2}} \erf(v) v^2 dv + \\
& & \int_{x=0}^\infty  \frac {x \sqrt{1 - c_0^2}}{\sqrt{2 \pi}} p(x)dx \cdot [- e^{-z^2/2}]_{-\frac{-c_0 x}{\sqrt{1 - c_0^2}}}^\infty \\
& = & \frac{c_0}{2} \int_{x=0}^\infty x^2 p(x) dx + \int_{v=0}^\infty \frac{(1 - c_0^2)^{1.5}}{c_0^2 \sqrt{\pi}} e^{-\frac{v^2 (1 - c_0^2}{c_0^2}} \erf(v) v^2 dv + \\
& & \int_{x=0}^\infty  \frac {x \sqrt{1 - c_0^2}}{\sqrt{2 \pi}} p(x)dx \cdot e^{-\frac{c_0^2 x^2}{2(1-c_0^2)}} \\
& = & \frac{c_0}{2} \frac{Var(x)}{2} + \int_{v=0}^\infty \frac{(1 - c_0^2)^{1.5}}{c_0^2 \sqrt{\pi}} e^{-\frac{v^2 (1 - c_0^2}{c_0^2}} \erf(v) v^2 dv + \\
& & \int_{x=0}^\infty  \frac {x \sqrt{1 - c_0^2}}{2 \pi} \cdot e^{-\frac{x^2}{2} - \frac{c_0^2 x^2}{2(1-c_0^2)}} dx \\
& = & \frac{c_0}{4} + \frac{(1 - c_0^2)^{1.5}}{c_0^2 \sqrt{\pi}} \int_{v=0}^\infty  e^{-\frac{v^2 (1 - c_0^2}{c_0^2}} \erf(v) v^2 dv + 
 \int_{x=0}^\infty  \frac {x \sqrt{1 - c_0^2}}{2 \pi} \cdot e^{-\frac{x^2}{2(1-c_0^2)}} dx \\
& = & \frac{c_0}{4} + \frac{(1 - c_0^2)^{1.5}}{c_0^2 \sqrt{\pi}} \int_{v=0}^\infty  e^{-\frac{v^2 (1 - c_0^2}{c_0^2}} \erf(v) v^2 dv + \int_{v=0}^\infty  \frac {v (1 - c_0^2)^{1.5}} {2 \pi} \cdot e^{-\frac{v^2}{2}} dv \\
& = & \frac{c_0}{4} + \frac{(1 - c_0^2)^{1.5}}{c_0^2 \sqrt{\pi}} \int_{v=0}^\infty  e^{-\frac{v^2 (1 - c_0^2}{c_0^2}} \erf(v) v^2 dv + \frac {(1 - c_0^2)^{1.5}} {2 \pi} 
\end{eqnarray*}
where in the second last step, we substituted $v = \frac{x}{\sqrt{1 - c_0^2}}$

We now use the well-known expression for the Error function integral (\cite{ng1969}) 
\begin{equation}
\int_0^\infty x^2 \erf(ax) e^{-b^2x^2} = \frac{\sqrt{\pi}}{4b^3} - \frac{\tan^{-1}{b/a}}{2 \sqrt{\pi} b^3} +  \frac{a}{2 \sqrt{\pi} b^2 (a^2 + b^2)}
\end{equation}

Thus, we have 
\begin{eqnarray*}
E_{x,y}[\relu(x)) \cdot \relu(y)] & = & 
\frac{c_0}{4} + \frac{(1 - c_0^2)^{1.5}}{c_0^2 \sqrt{\pi}}(\frac{\sqrt{\pi} c_0^3}{4(1-c_0^2)^{1.5}} - \frac{\tan^{-1}{\sqrt{\frac{1-c_0^2}{c_0^2}}} c_0^3}{2 \sqrt{\pi} (1-c_0^2)^{1.5}} +  \frac{c_0^4}{2 \sqrt{\pi} (1 - c_0^2)}) + \frac {(1 - c_0^2)^{1.5}} {2 \pi} \\
& = & \frac{c_0}{2} - \frac{c_0}{2\pi} \tan^{-1}{\sqrt{\frac{1-c_0^2}{c_0^2}}} + \frac{c_0^2 \sqrt{1 - c_0^2}}{2\pi} + \frac {(1 - c_0^2)^{1.5}} {2 \pi} \\
& = & \frac{c_0}{2} - \frac{c_0}{2\pi} \tan^{-1}{\sqrt{\frac{1-c_0^2}{c_0^2}}} + \frac{\sqrt{1 - c_0^2}}{2\pi}
\end{eqnarray*}
Now, $E[B(\relu(x)) \cdot B(\relu(y))] = E[\frac{1}{s^2}(\relu(x) - m)(\relu(y) - m]$,
where $m=\mean_{Z \in \mathcal{N}(0,1)}(\relu(Z))$ and $s = \std_{Z \in \mathcal{N}(0,1)}(\relu(Z))$. We now compute $m$ and $s^2$.

\begin{eqnarray*}
m & = &\int_{x=0}^\infty \frac{x}{\sqrt{2\pi}} e^{-x^2/2}dx \\
& = & \frac{1}{\sqrt{2\pi}}[- e^{-z^2/2}]_0^\infty = \frac{1}{\sqrt{2\pi}} 
\end{eqnarray*}

\begin{eqnarray*}
s^2 & = &\int_{x=0}^\infty \frac{x^2}{\sqrt{2\pi}} e^{-x^2/2}dx - m^2\\
& = & \frac{1}{2}\int_{x=-\infty}^\infty \frac{x^2}{\sqrt{2\pi}} e^{-x^2/2}dx - m^2 = \frac{1}{2} - \frac{1}{2\pi} 
\end{eqnarray*}

Finally, we obtain 
\begin{eqnarray*}
\frac{E[B(\relu(x)) \cdot B(\relu(y))]}{c_0} & = & E[\frac{1}{s^2 c_0}(\relu(x) - m)(\relu(y) - m]  = \frac{1}{s^2 c_0}(E[\relu(x)\cdot \relu(y)]) - m^2)\\
& = & \frac{1}{s^2 c_0}(\frac{c_0}{2} - \frac{c_0}{2\pi} \tan^{-1}{\sqrt{\frac{1-c_0^2}{c_0^2}}} + \frac{\sqrt{1 - c_0^2}}{2\pi} - \frac{1}{2\pi}) \\
& = & \frac{1}{s^2}(\frac{1}{2} - \frac{1}{2\pi} \tan^{-1}{\sqrt{\frac{1-c_0^2}{c_0^2}}} + \frac{\sqrt{1 - c_0^2}}{2\pi} - \frac{1}{2\pi c_0})  \\
& = & \frac{1}{\pi - 1}(\pi - \tan^{-1}{\sqrt{\frac{1-c_0^2}{c_0^2}}} + \frac{\sqrt{1 - c_0^2}}{c_0} - \frac{1}{c_0})  \\
& = & \frac{\pi - (\theta - \tan{\theta} + \frac{1}{\cos{\theta}})}{\pi -1}  \leq 1 \\
\end{eqnarray*}
where we used $\theta =\cos^{-1}{c_0}$

It can be seen numerically that the function $\theta - \sin{\theta} + \frac{1}{\cos{\theta}}$ is lower bounded by $1$, and it attains this minimum value at $\theta=0$.
\end{proof}

\bibliographystyle{abbrv}
\bibliography{learnability}

\end{document}